\documentclass[11pt]{article}

\usepackage[preprint]{acl}

\usepackage{times}
\usepackage{latexsym}

\usepackage[utf8]{inputenc}

\usepackage{CJKutf8}

\usepackage{microtype}

\usepackage{inconsolata}
\usepackage{fvextra}

\usepackage{graphicx}
\usepackage{subcaption}
\usepackage{xcolor}
\usepackage[most]{tcolorbox}
\usepackage{booktabs}
\usepackage{multirow}
\usepackage{amsmath}
\usepackage{amssymb}

\title{When Does Generating More Help? Disentangling Fixed-Source Synthesis from Source Expansion in Synthetic Data Scaling}

\definecolor{figBlue}{HTML}{3B6FCB}
\definecolor{promptBlue}{RGB}{0,102,204}

\author{
  \textbf{Xu Guo}$^{1,2,3}$ \quad
  \textbf{Jian Tong}$^{1}$ \quad
  \textbf{Zhihui Lu}$^{\dagger 3}$ \quad
  \textbf{Qipeng Guo}$^{\dagger 1,2}$ \\[6pt]
  $^{1}$Shanghai AI Laboratory \quad
  $^{2}$Shanghai Innovation Institute \quad
  $^{3}$Fudan University \\[4pt]
  {\small \texttt{\{tongjian, guoqipeng\}@pjlab.org.cn}} \\
  {\small \texttt{guox24@m.fudan.edu.cn} \quad \texttt{lzh@fudan.edu.cn}} \\[2pt]
  {\small $^\dagger$Corresponding authors}
}

\begin{document}
\begin{CJK}{UTF8}{gbsn}  

\maketitle
\begin{abstract}
Synthetic data can be scaled along two routes: \emph{Source Expansion} (SE), which enlarges the source by adding seed materials or generators, 
and \emph{Fixed-Source Synthesis} (FSS), which holds the source fixed and scales the generation budget. 
Existing scaling studies typically expand the source as the data grows, conflating SE with FSS and leaving FSS underexplored. 
We isolate FSS by holding the seed-question pool and teacher model fixed, varying only the per-question response budget under Rejection Sampling (RS). 
We adapt the rectified scaling law to FSS, deriving it from how repeated sampling covers a fixed source. 
Empirically, the derived form, fit on low budgets, predicts performance at the held-out highest budget for every evaluated teacher--student pair.
At matched total-sample budgets, SE and FSS are comparable at small budgets; at large budgets, adding seed questions outperforms spending the same budget on more responses. 
Within FSS, however, neither synthesizing additional questions from the existing seeds nor varying the synthesis protocol outperforms plain RS at matched budgets.
FSS is thus a bounded scaling axis and a controlled setting for comparing synthesis protocols.
We will release our code and data to facilitate further research.
\end{abstract}

\section{Introduction}
\label{sec:intro}

Synthetic data is now standard in training large language models (LLMs)~\citep{wang2023selfinstruct, mukherjee2023orca, xu2023wizardlm, toshniwal2024openmathinstruct2, yue2024mammoth2}.
Prior studies report that downstream performance scales predictably with synthetic datasets~\citep{ge2024-scaling-synthetic-data-personas,yue2024mammoth2,busbridge2025-distillation-scaling-laws}.

\begin{figure}[t]
    \centering
    \includegraphics[width=\linewidth]{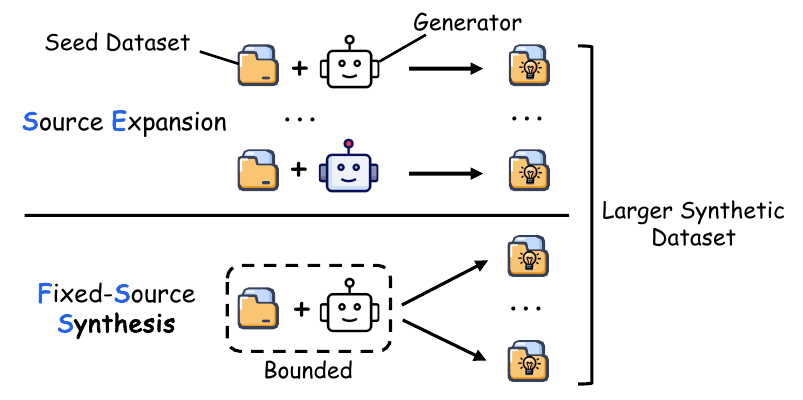}
    \caption{Two routes for scaling synthetic data. \textcolor{figBlue}{\textbf{S}}ource \textcolor{figBlue}{\textbf{E}}xpansion (\textcolor{figBlue}{\textbf{SE}}) enlarges the information sources (the seed set and generator), while \textcolor{figBlue}{\textbf{F}}ixed-\textcolor{figBlue}{\textbf{S}}ource \textcolor{figBlue}{\textbf{S}}ynthesis (\textcolor{figBlue}{\textbf{FSS}}) holds the source fixed and scales only the generation budget.}
    \label{fig:intro-concept}
\end{figure}

However, synthetic data scales along two distinct routes. A synthetic dataset can grow either by enlarging the source (adding documents, seed questions, or generators), or by holding the source fixed and scaling the generation budget.
We refer to these as \emph{source expansion} (SE) and \emph{fixed-source synthesis} (FSS), respectively (Fig.~\ref{fig:intro-concept}).

Recent studies scale along the SE route: \citet{yue2024mammoth2} retrieve more web documents and extract a few instructions per document, and \citet{qin2025scaling} expand the source-document pool for math while holding per-document generation fixed.
These are natural practices for building corpora, but they scale along SE, leaving FSS returns underexplored.

We isolate FSS within a standard synthesis pipeline: the source is a pair $(Q, T)$ of a seed-question pool $Q$ and a teacher $T$; FSS fixes $(Q, T)$ and scales the per-question response budget $r$ via rejection sampling (RS), while SE enlarges $Q$ or switches $T$. Within this setup, we ask three research questions:

$\circ$ \textbf{RQ1.} With the source $(Q,T)$ fixed, does FSS improve student accuracy, and how does the gain scale with $r$?

$\circ$ \textbf{RQ2.} At matched total-sample budgets, how do SE and FSS gains compare?

$\circ$ \textbf{RQ3.} Within FSS, do common synthesis protocols (diversifying, selecting, or rewriting the synthetic data) improve over plain RS?

For RQ1, we scale $r$ under RS and trace student accuracy on two domains (Mathematics, Physics).
For RQ2, we run two comparisons at a fixed total-sample budget: expanding the real seed pool $Q$ (SE) versus increasing $r$ at fixed $(Q,T)$ (FSS); and synthesizing additional questions from the existing seeds versus plain RS.
For RQ3, we evaluate common synthesis protocols (alternative sampling temperatures, embedding-based selection, persona prompting, unfiltered sampling, and trace-level repair) against plain RS under the same fixed-source control.
In total, synthetic-data construction consumes about $2$B teacher-inference tokens ($1.6$B from DeepSeek-V3.1 rollouts), followed by over $500$ student SFT runs (more than $2{,}000$ GPU-hours of fine-tuning).
Below, we summarize the key findings for each RQ.

$\circ$ FSS gains diminish with $r$ and leave a clear teacher--student gap at the largest budget; a fixed-source form, derived from repeated sampling and fit on low budgets, predicts performance at the held-out highest budget for every evaluated teacher--student pair.

$\circ$ At matched total-sample budgets, SE and FSS are comparable at small budgets, but expanding the real seed-question set (SE) becomes the more reliable axis as the budget grows; synthesizing additional questions from the existing seeds yields no gain over plain RS.

$\circ$ Under the fixed-source control, common synthesis protocols do not consistently improve over plain RS, so protocol $p$ contributes little once the source $(Q,T)$ is fixed.

\section{Related Work}
\label{sec:related}

\paragraph{Scaling laws for synthetic-data fine-tuning.}
Scaling laws characterize how pre-training loss varies with model size and training tokens, providing a principled basis for compute allocation~\citep{hestness2017deep, kaplanscaling, chinchilla}.
Subsequent studies extend this view beyond pre-training tokens, covering data-constrained scaling under limited unique tokens~\citep{muennighoff2023scaling, lovelace2026prescriptivescalinglawsdata} and the role of teacher strength in distillation~\citep{bansal2024-smaller-weaker-yet-better}.
A closer sub-line targets fine-tuning: \citet{hernandez2021scaling} characterize transfer from pre-training, \citet{lin2024selecting} propose a rectified scaling law for fine-tuning, and \citet{qin2025scaling} apply this rectified law to synthetic-data fine-tuning. 
These studies scale synthetic data by enlarging the source.
In contrast, we fix the source $(Q,T)$ and trace SFT scaling within FSS alone.

\paragraph{Synthetic Data Generation.}
A second line of work targets the synthesis protocol $p$.
Prompt-synthesis methods generate new prompts from the seed set~\citep{wang2023selfinstruct,xu2023wizardlm,yue2024mammoth2}.
Sampling-side methods alter teacher-trace generation, either by reallocating generation budget across teachers or samples~\citep{bansal2024-smaller-weaker-yet-better} or through prompt-side conditioning such as persona-driven generation~\citep{ge2024-scaling-synthetic-data-personas}. Data curation methods act on the resulting candidates: embedding- or gradient-based selection~\citep{zhang2024-tagcos, liu2023-what-makes-good-data, xia2024-less} retains compact, high-value subsets, while filtering controls how much teacher noise enters training, including settings that omit filtering or use negative reasoning traces as supervision~\citep{mukherjee2023orca,xu2025harnessing-negative-signals}.  Critique-and-revise methods rewrite weak or incorrect traces~\citep{madaan2023-self-refine, kapusuzoglu2025-critique-guided-distillation, yang2024-supercorrect}.
These studies, however, evaluate protocols across heterogeneous teachers and seed corpora, entangling the effect of $p$ with that of $(Q,T)$. We therefore re-evaluate representative protocols per strand in a teacher--student setting with $(Q,T)$ fixed, so that gains reflect $p$ alone.

\section{Source Expansion vs.\ Fixed-Source Synthesis}
\label{sec:prelim}

We model a synthetic-data pipeline as a triple $(Q, T, p)$: a seed-question set $Q$, a teacher $T$, and a synthesis protocol $p$ that turns seeds into training examples through choices of prompt template, decoding parameters, filtering, selection, or rewriting.
For each seed $q \in Q$, $p$ generates one or more SFT examples from $(q, T)$; the training set is the multiset of all such examples.
We distinguish two routes for enlarging this set:

$\circ$ \textbf{Source expansion (SE)} alters $(Q, T)$ by adding real seed questions from outside the pool, or switching to a stronger or additional teacher.

$\circ$ \textbf{Fixed-source synthesis (FSS)} fixes the source $(Q, T)$ and varies only the protocol $p$ or the per-question budget $r$, including protocols that synthesize new prompts from the existing seeds.

The three RQs partition this triple: Sec.~\ref{sec:rq1} fixes $(Q,T)$ and sweeps the per-question budget $r$ (FSS); Sec.~\ref{sec:rq2} matches total-sample budgets between SE (scaling $|Q|$) and FSS (scaling $r$); Sec.~\ref{sec:rq3} fixes $(Q,T)$ and varies the synthesis protocol $p$.

\section{Experimental Setup}
\label{sec:setup}

The setup below is shared across all three RQs; implementation details appear in Appendix~\ref{sec:appendix}.

\paragraph{Domains.}
We study two reasoning domains from SuperGPQA~\citep{du2025-supergpqa}: \textbf{Mathematics} and \textbf{Physics}. Tab.~\ref{tab:domains} reports train/test sizes and per-domain teacher references.

\paragraph{Models.}
The primary teacher is DeepSeek-V3.1 (\texttt{dv31}; $T_{\mathrm{DS}}$), and the primary student is \texttt{Qwen2-7B-Instruct} ($S_{\mathrm{Q7}}$).  We use a second teacher (\texttt{Qwen2.5-72B-Instruct}; $T_{\mathrm{Q72}}$) and a second student (\texttt{Llama-3.1-8B-Instruct}; $S_{\mathrm{L8}}$) for the cross-pair RQ1 checks and fixed-source audit in Appendix~\ref{sec:appendix-cross-pair}. The heterogeneous-source check in Sec.~\ref{sec:mixed-stem} uses \texttt{Qwen3-4B} ($S_{\mathrm{Q3}}$) as the student.

\paragraph{Training.}
Each student is fine-tuned by SFT for the same number of epochs. For cells with repeated SFT runs, we use $n{=}3$ training seeds and report $\mathrm{mean} \pm \mathrm{std}$.

\paragraph{Evaluation.}
The primary metric is $\mathrm{mean@8}$: for each test question, we sample $8$ student responses at temperature $0.7$ and report the mean fraction correct. Averaging over $8$ samples reduces per-question sampling noise and stabilizes the comparisons below.

\paragraph{Baseline protocol.}
Our baseline protocol $p_{\text{RS}}$ is RS. For each $q \in Q$, we sample teacher responses at temperature $1.2$ until $32$ correct ones are collected; at training budget $r \in \{1,2,4,8,16,32\}$, we use the first $r$ as the per-question training set.
Alternative protocols (Sec.~\ref{sec:rq3}) modify only $p$ (temperature, selection criterion, prompting style, filter, or trace structure) while keeping $(Q, T)$ fixed. We study SE, which instead scales $|Q|$, in Sec.~\ref{sec:rq2}.

\begin{table}[t]
    \centering
    \small
    \setlength{\tabcolsep}{3.5pt}
    \begin{tabular}{@{}lrrcccc@{}}
        \toprule
        Domain & $|Q_{\text{tr}}|$ & $|Q_{\text{te}}|$ & $T_{\mathrm{DS}}$ & $T_{\mathrm{Q72}}$ & $S_{\mathrm{Q7}}$ & $S_{\mathrm{L8}}$ \\
        \midrule
        Mathematics & $1833$ & $406$ & $0.753$ & $0.466$ & $0.221$ & $0.174$ \\
        Physics     & $2010$ & $458$ & $0.605$ & $0.365$ & $0.181$ & $0.161$ \\
        \bottomrule
    \end{tabular}
    \caption{Per-domain train/test sizes and reference $\mathrm{mean@8}$ scores. Model abbreviations follow Sec.~\ref{sec:setup}.}
    \label{tab:domains}
\end{table}

\begin{table*}[t]
    \centering
    \small
    \setlength{\tabcolsep}{4pt}
    \begin{tabular}{llccccccc}
        \toprule
        Domain & Pair & $r{=}1$ & $r{=}2$ & $r{=}4$ & $r{=}8$ & $r{=}16$ & $r{=}32$ & best $r$ \\
        \midrule
        \multirow{4}{*}{Mathematics}
            & $T_{\mathrm{DS}}{\to}S_{\mathrm{Q7}}$  & $0.317_{\pm 0.009}$ & $0.330_{\pm 0.003}$ & $0.357_{\pm 0.004}$ & $0.381_{\pm 0.006}$ & $0.398_{\pm 0.007}$ & $\mathbf{0.424}_{\pm 0.005}$ & $32$ \\
            & $T_{\mathrm{DS}}{\to}S_{\mathrm{L8}}$  & $0.293_{\pm 0.008}$ & $0.318_{\pm 0.007}$ & $0.324_{\pm 0.012}$ & $0.347_{\pm 0.006}$ & $0.367_{\pm 0.014}$ & $\mathbf{0.386}_{\pm 0.006}$ & $32$ \\
            & $T_{\mathrm{Q72}}{\to}S_{\mathrm{Q7}}$ & $0.307_{\pm 0.003}$ & $0.312_{\pm 0.003}$ & $0.322_{\pm 0.002}$ & $0.330_{\pm 0.008}$ & $0.329_{\pm 0.006}$ & $\mathbf{0.338}_{\pm 0.004}$ & $32$ \\
            & $T_{\mathrm{Q72}}{\to}S_{\mathrm{L8}}$ & $0.296_{\pm 0.012}$ & $0.317_{\pm 0.010}$ & $0.320_{\pm 0.009}$ & $0.338_{\pm 0.003}$ & $\mathbf{0.341}_{\pm 0.003}$ & $0.340_{\pm 0.005}$ & $16$ \\
        \midrule
        \multirow{4}{*}{Physics}
            & $T_{\mathrm{DS}}{\to}S_{\mathrm{Q7}}$  & $0.222_{\pm 0.004}$ & $0.236_{\pm 0.003}$ & $0.250_{\pm 0.002}$ & $0.266_{\pm 0.008}$ & $0.285_{\pm 0.007}$ & $\mathbf{0.302}_{\pm 0.009}$ & $32$ \\
            & $T_{\mathrm{DS}}{\to}S_{\mathrm{L8}}$  & $0.251_{\pm 0.005}$ & $0.260_{\pm 0.004}$ & $0.265_{\pm 0.004}$ & $0.291_{\pm 0.005}$ & $0.311_{\pm 0.003}$ & $\mathbf{0.327}_{\pm 0.004}$ & $32$ \\
            & $T_{\mathrm{Q72}}{\to}S_{\mathrm{Q7}}$ & $0.215_{\pm 0.005}$ & $0.226_{\pm 0.006}$ & $0.243_{\pm 0.002}$ & $0.251_{\pm 0.002}$ & $0.256_{\pm 0.006}$ & $\mathbf{0.262}_{\pm 0.005}$ & $32$ \\
            & $T_{\mathrm{Q72}}{\to}S_{\mathrm{L8}}$ & $0.228_{\pm 0.004}$ & $0.246_{\pm 0.002}$ & $0.262_{\pm 0.004}$ & $0.263_{\pm 0.003}$ & $\mathbf{0.269}_{\pm 0.001}$ & $0.267_{\pm 0.003}$ & $16$ \\
        \bottomrule
    \end{tabular}
    \caption{Student $\mathrm{mean@8}$ versus RS budget $r$ for the four teacher--student pairs on Mathematics and Physics. Cells report $\mathrm{mean}_{\pm\mathrm{std}}$ over $3$ train seeds; bold marks the best-$r$ cell in each row.}
    \label{tab:rq1}
\end{table*}

\section{RQ1: Fixed-Source Synthesis Scaling}
\label{sec:rq1}

This section quantifies how far FSS alone can raise student performance under a fixed source $(Q, T)$. We define the source-determined ceiling that bounds FSS, derive a three-parameter form for the response-budget curve $P(r)$, and identify the budget at which the empirical curve flattens.

\subsection{Response-scaling form for FSS}

A common parametric form for fine-tuning data scaling is the \emph{rectified scaling law}~\citep{lin2024selecting},
\begin{equation}
L(D) \;=\; \frac{B}{D_\ell + D^{\beta}} + E,
\label{eq:rectified}
\end{equation}
where $D$ is the fine-tuning data size, $L(D)$ the validation loss, $E$ the irreducible error as $D \to \infty$, $B>0$ an amplitude, $\beta>0$ the scaling exponent, and $D_\ell>0$ a rectification constant capturing the model's pre-learned competence.
\citet{qin2025scaling} apply this form to synthetic data with $L(D)$ as the test error rate. Substituting the test accuracy $P = 1 - L$ into Eq.~\ref{eq:rectified} gives
\begin{equation}
P(D) \;=\; (1 - E) \;-\; \frac{B}{D_\ell + D^{\beta}},
\label{eq:rectified-acc}
\end{equation}
i.e.\ a fixed ceiling $1-E$ minus a term that vanishes as $D \to \infty$. In the rectified-law reading, $1-E$ is the asymptote of the underlying prediction task: an irreducible residual that no additional data removes, set by the task's intrinsic noise.

FSS differs in the \emph{origin} of its bound. The scaling axis is no longer data but a budget spent on a fixed source: with $(Q, T)$ held fixed and only the per-question response budget $r$ growing, each additional draw rephrases or diversifies the same fixed content. 
The relevant asymptote is therefore different. It is no longer the prediction task's irreducible-error ceiling $1-E$, which assumes an unbounded ideal data source, but a \emph{source}-determined ceiling $P_\infty(Q, T)$: the best test performance reachable while $(Q, T)$ is held fixed.
We retain the parametric family of Eq.~\ref{eq:rectified-acc} and re-derive its functional form under this fixed-source view.

To derive the shape of $P(r)$, we model the source through discrete latent features $f$, following the \emph{quanta} view of \citet{michaud2023quantization}: individually learnable units of skill or reasoning that a student either acquires or does not.
Coverage is \emph{binary}: a feature contributes if at least one training sample exposes it, with no partial credit.
Let $q(f)$ denote its per-sample exposure probability. Under i.i.d.\ draws, $f$ remains uncovered after $r$ samples with probability $(1-q(f))^r$, giving $P(r) = P_\infty - \mathbb{E}_f[(1-q(f))^r]$.
Our distributional assumption is that $q(f)$ follows a long-tailed (power-law) distribution over features, as predicted by the same quanta view.

We then split the expectation $\mathbb{E}_f[(1-q(f))^r]$ by feature frequency. Common features are covered within a few draws, with miss probability decaying exponentially in $r$; once $r$ is moderate, the \emph{rare}-feature tail dominates the residual gap to $P_\infty$.
Under a power-law tail, this residual decays polynomially in $r$, giving
\begin{equation}
P(r) \;\approx\; P_\infty - A\,r^{-\alpha},
\label{eq:rq1-hypothesis}
\end{equation}
the offset-free ($D_\ell{=}0$) case of Eq.~\ref{eq:rectified-acc} with source-determined ceiling $P_\infty$.
Performance thus rises quickly while common features are covered, then approaches $P_\infty$ at the polynomial rate $r^{-\alpha}$. The full derivation is deferred to Appendix~\ref{sec:appendix-derivation}.

\subsection{Experiments for RQ1}

\begin{figure*}[t]
    \centering
    \includegraphics[width=0.95\textwidth]{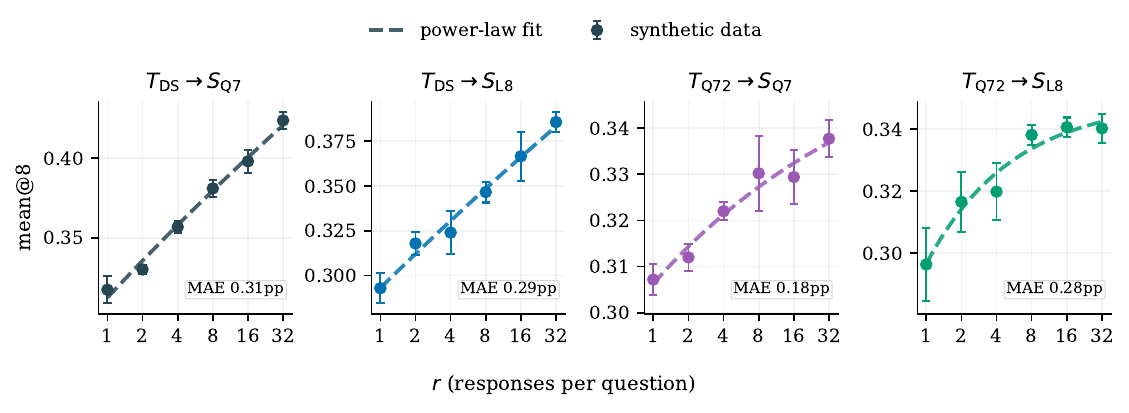}
    \caption{Mathematics RS response-scaling. A single parametric form fits all four teacher--student pairs. Error bars are $\pm$std over $n{=}3$ train seeds. MAE denotes mean absolute error between observed and fitted performance.}
    \label{fig:rq1-cross-pair}
\end{figure*}

\begin{figure*}[t]
    \centering
    \includegraphics[width=0.95\textwidth]{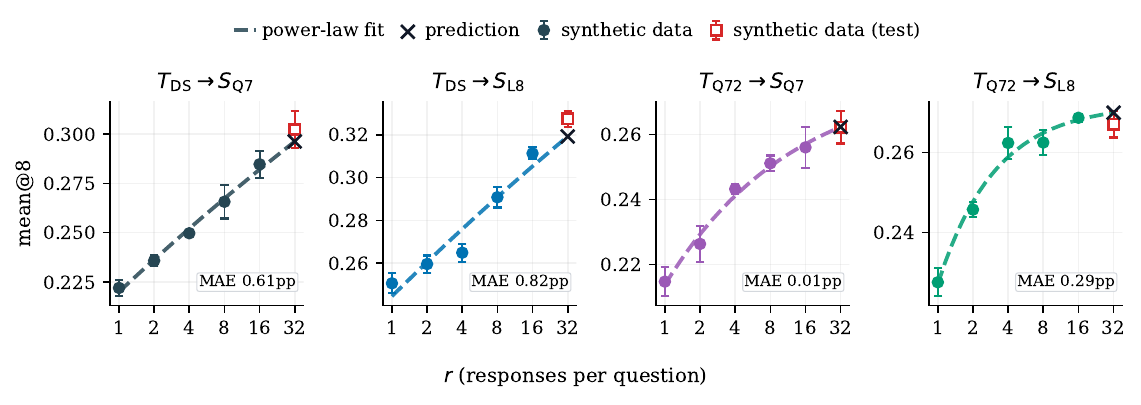}
    \caption{Physics RS response-scaling. We fit on $r\le 16$ and forward-predict $r{=}32$, then compare to the observed point. Error bars are $\pm$std over $n{=}3$ train seeds.}
    \label{fig:rq1-forecast}
\end{figure*}

We hold $(Q, T)$ fixed and sweep the RS budget $r$ across four teacher--student pairs ($T_{\mathrm{DS}}{\to}S_{\mathrm{Q7}}$, $T_{\mathrm{DS}}{\to}S_{\mathrm{L8}}$, $T_{\mathrm{Q72}}{\to}S_{\mathrm{Q7}}$, $T_{\mathrm{Q72}}{\to}S_{\mathrm{L8}}$), using $n{=}3$ training seeds per configuration.
Tab.~\ref{tab:rq1} reports $\mathrm{mean@8}$ on Mathematics and Physics across the four pairs.
We subject Eq.~\ref{eq:rq1-hypothesis} to two complementary tests.
On Mathematics, we fit to the full $r$ grid and test whether a single parametric form fits all four teacher--student pairs (Fig.~\ref{fig:rq1-cross-pair}).
On Physics, we test held-out forecasting: we fit the same form on $r \le 16$ and predict $r = 32$ for each pair (Fig.~\ref{fig:rq1-forecast}), following standard practice in fine-tuning scaling-law studies~\citep{lin2024selecting,qin2025scaling}.
Appendix~\ref{sec:appendix-form} runs the same held-out forecast on all four pairs in both domains.
We report fit quality as the mean absolute error (in percentage points) over an evaluation grid $\mathcal{R}$:
\begin{equation}
\mathrm{MAE}_{\mathrm{pp}} = \frac{100}{|\mathcal{R}|}\!\sum_{r\in\mathcal{R}}\! \bigl|P_{\mathrm{obs}}(r)-P_{\mathrm{fit}}(r)\bigr|.
\label{eq:mae-pp}
\end{equation}
For Mathematics, $\mathcal{R}$ is the full $r$ grid (in-sample fit); for Physics, $\mathcal{R}=\{32\}$ (held-out forecast error).
Appendix~\ref{sec:appendix-form} reports a sanity check against trivial extrapolation baselines.

\subsection{Analysis for RQ1}

 Tab.~\ref{tab:rq1} and Figs.~\ref{fig:rq1-cross-pair}--\ref{fig:rq1-forecast} report the results, from which we draw two observations.

\paragraph{(i) FSS shows diminishing returns in the response budget.}
Across Mathematics, Physics, and all four teacher--student pairs, student accuracy keeps rising with $r$. The budget axis is logarithmic, so a steadily climbing curve corresponds to a shrinking gain per additional response.
This is consistent with the saturating form of Eq.~\ref{eq:rq1-hypothesis}: with $(Q,T)$ held fixed, early responses cover the common features, while later responses increasingly repeat already-covered signal and reach the rare tail only occasionally, so the marginal gain per response shrinks.

\paragraph{(ii) A stronger teacher raises the source ceiling.}
Although diminishing returns appear in all settings, the ceiling depends on the teacher. 
Within the scanned grid, $T_{\mathrm{Q72}}$ begins to flatten.
On the same seed set and student, $T_{\mathrm{DS}}$ surpasses $T_{\mathrm{Q72}}$ and keeps rising.
The stronger teacher thus lifts the source ceiling above the level the weaker teacher can reach.

\paragraph{Summary.}
RQ1 supports the fixed-source scaling hypothesis: FSS consistently improves student accuracy, but its marginal gains diminish as the response budget $r$ grows.
Saturation sets in earlier under the weaker teacher and later under the stronger one, so the benefit of an FSS depends jointly on $r$ and teacher strength.

\begin{figure*}[t]
    \centering
    \includegraphics[width=0.92\textwidth]{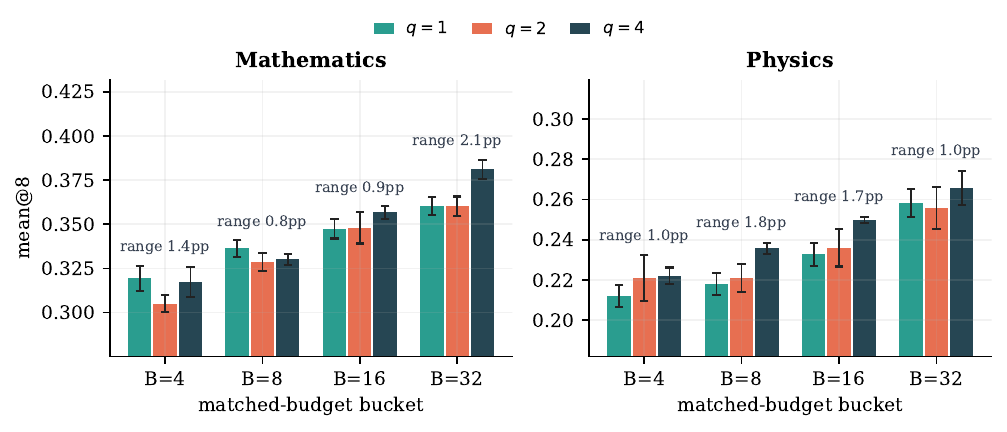}
    \caption{Matched-budget comparison between real-seed SE and response-budget FSS on Mathematics (left) and Physics (right). Within each domain, SE varies the real seed-question subset ($q{\in}\{1,2,4\}$) and FSS varies the per-question response count $r$. Each bucket fixes the total SFT-example count by grouping the three allocations that share the same product $B{\equiv}q\cdot r$. The annotated ``range'' is the spread across these three allocations within a bucket. Error bars show the standard deviation over $n{=}3$ train seeds in both domains.}
    \label{fig:rq2-iso-sample}
\end{figure*}

\begin{figure}[t]
    \centering
    \includegraphics[width=\columnwidth]{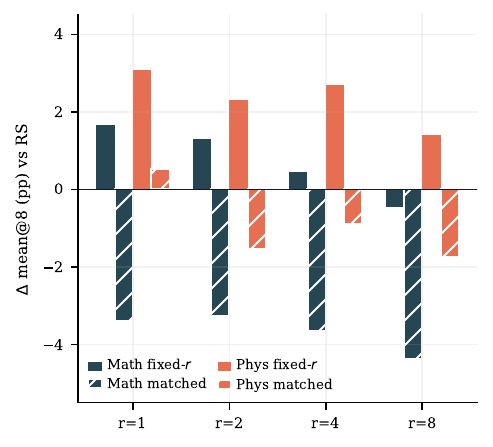}
    \caption{Question synthesis under fixed-response and matched-budget baselines. Solid bars: fixed-$r$ comparison. Hatched bars: matched-budget comparison.}
    \label{fig:question-synthesis}
\end{figure}

\section{RQ2: Matched-Budget SE and FSS Gains}
\label{sec:rq2}

This section asks two matched-budget questions. First, when we add real seed questions, how does SE compare with spending the same budget on more responses from fewer seeds? Second, when we synthesize new prompts from existing seeds, do they outperform an RS baseline?

\subsection{Experiments for RQ2}

\paragraph{Paradigm 1: add new seed questions.}
Let $Q_4$ denote the full real seed-question pool. Let $Q_1 \subset Q_2 \subset Q_4$ be nested random subsets holding one-quarter and one-half of its questions, respectively, so that $|Q_q| = \tfrac{q}{4}|Q_4|$.
We sweep the RS response budget $r$ and compare allocations matched on the number of SFT examples. An allocation $(q, r)$ yields $|Q_q|\cdot r = \tfrac{q}{4}|Q_4|\cdot r$ examples, so the budget depends only on the product $q\cdot r$. A smaller question pool therefore matches the full pool only by raising $r$ in proportion.
We index a matched-budget bucket by this product, $B \equiv q\cdot r \in \{4,8,16,32\}$: bucket $B$ collects $(q{=}4,\, r{=}B/4)$, $(q{=}2,\, r{=}B/2)$, and $(q{=}1,\, r{=}B)$. At $B{=}4$, for instance, $Q_4$ draws one response per question while $Q_1$ draws four.
As $B$ grows, the smaller-$|Q|$ allocations spend more responses on fewer seeds. We therefore predict that the larger-$|Q|$ allocation wins at larger $B$, because repeated sampling from a smaller seed pool should yield diminishing returns sooner. We run this matched-budget triple on both Mathematics and Physics, so the prediction is tested cross-domain.

\paragraph{Paradigm 2: question synthesis as FSS.}
We synthesize additional prompts from the full real seed set for Mathematics and Physics using an instruction-evolution procedure~\citep{xu2023wizardlm}, with our primary teacher $T_{\mathrm{DS}}$ (DeepSeek-V3.1) as the synthesis model;
format and similarity filtering leave $\sim\!4\times$ as many prompts as the seed set, on which we sweep $r \in \{1,2,4,8\}$ under the same RS protocol.
Because these prompts derive from the existing seeds rather than new source material, question synthesis stays within the fixed source.
We compare against the RS baseline under two references (Fig.~\ref{fig:question-synthesis}): the \emph{fixed-$r$} reference compares the synthesized-question condition at $r{=}k$ with RS on the original seeds at the same $r{=}k$, so the synthesized side contributes more SFT examples;
the \emph{matched-budget} reference compares it with RS at $r{=}4k$, holding the total SFT examples equal so the comparison isolates how the synthesized prompts spend the fixed-source budget.

\begin{figure*}[!t]
    \centering
    \includegraphics[width=\textwidth]{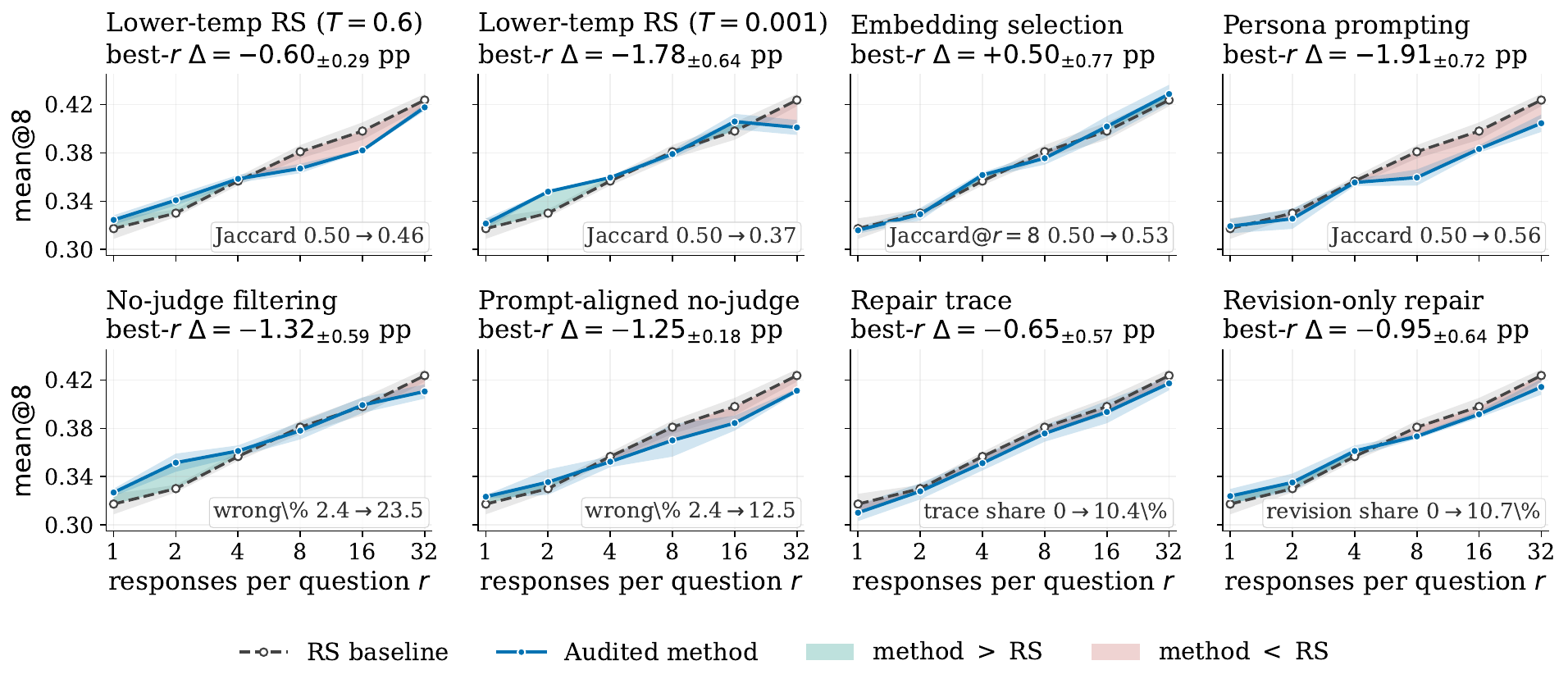}
    \caption{RQ3 audit on Mathematics. Each panel overlays one audited protocol's $r$-sweep (blue, averaged over $n{=}3$ training seeds) on the shared RS baseline (dashed line). Bands show the train-seed standard deviation.
    Shading between the protocol and RS curves is blue where the protocol exceeds the baseline and red where it falls below.
    The bottom-right annotation reports the shift in the intrinsic metric from the RS reference to the audited pool (Jaccard distance for diversity protocols, wrong-response \% for filter ablations, rewrite share for trace-rewrite variants).
    Each protocol shifts its targeted intrinsic metric, yet none yields a stable gain at its best $r$.}
    \label{fig:rq3-panels}
\end{figure*}

\subsection{Analysis for RQ2}

Figs.~\ref{fig:rq2-iso-sample} and~\ref{fig:question-synthesis} address one question: under a fixed budget, where should the next training example go? 
We compare three uses of that example: a new real seed question, an additional response to an existing seed, and a question synthesized from existing seeds.
We have the following two findings.

\paragraph{(i) The next example is better spent on a new real seed than on more responses to existing seeds.}
On Physics, $q{=}4$ is the best allocation at every budget; on Mathematics, $q{=}4$ matches $q{=}1$ at low budget and overtakes it at high budget (Fig.~\ref{fig:rq2-iso-sample}).
This is consistent with the fixed-source account: extra responses to a few seeds exhaust their limited source signal, whereas new seed questions add coverage that resampling existing seeds cannot supply.

\paragraph{(ii) When no new real seed is available, the next example is better spent on another real response than on a synthesized question.}
Under the fixed-$r$ reference, synthesized questions beat the same-$r$ RS baseline (Fig.~\ref{fig:question-synthesis}, solid bars).
Under the matched-budget reference the comparison reverses: once the total SFT-example count is equalized, more responses from the original seeds beat synthesized questions (Fig.~\ref{fig:question-synthesis}, hatched bars).
This result is specific to a fixed synthesis budget and to questions directly conditioned on the existing seeds.
A larger synthesis budget, or a different downstream task setting, may still make synthesized questions worthwhile.

\paragraph{Summary.}
RQ2 asks how to allocate a fixed budget across data sources.
The matched-budget comparison isolates the cause: the apparent payoff of prompt synthesis is a budget effect, not a synthesis effect. 
Once the example count is held equal, prompts synthesized from existing seeds underperform plain re-sampling of the original seeds.

\section{RQ3: Synthesis Protocols at Fixed Budget}
\label{sec:rq3}

Under fixed $(Q,T)$, we ask whether any synthesis protocol improves over the RS reference.
Fig.~\ref{fig:rq3-panels} overlays each audited protocol's $r$-sweep on the shared RS baseline.

\subsection{Experiments for RQ3}

The audited methods are listed below. 
All audited protocols and their implementation details are provided in Appendix~\ref{sec:appendix}.

$\circ$ \textbf{Temperature variants}: replace the RS sampling temperature ($1.2$) with $0.6$ or $0.001$.

$\circ$ \textbf{Embedding-based selection}: from the RS pool, pick $r$ responses per question to maximize pairwise embedding distance, following the diversity criterion of DEITA~\citep{liu2023-what-makes-good-data}; the $r{=}1$ and $r{=}32$ endpoints match RS by construction.

$\circ$ \textbf{Persona prompting}: prepend a persona sampled from PersonaHub~\citep{ge2024-scaling-synthetic-data-personas} as a system message before each teacher generation.

$\circ$ \textbf{Judge-free / filter ablation}: drop the correctness judge to admit wrong teacher traces, connecting to work that uses negative reasoning as supervision~\citep{mukherjee2023orca,xu2025harnessing-negative-signals}; a prompt-aligned variant pins the prompt set to the RS subset, isolating response filtering from question coverage.

$\circ$ \textbf{Trace-level repair}: rewrite wrong teacher responses into a \texttt{wrong} $\to$ \texttt{check} $\to$ \texttt{fix} trace, adapting iterative-refinement and critique-then-revise recipes~\citep{madaan2023-self-refine,yang2024-supercorrect,kapusuzoglu2025-critique-guided-distillation}; a second variant keeps only the revised correct solution.

\paragraph{Intrinsic metrics.}
For each protocol, we pair the transfer outcome with an \emph{intrinsic} scalar of the audited SFT pool that should shift when the intervention takes effect.
\textit{Response diversity}, the mean pairwise token-set Jaccard distance across the retained responses per question, tracks the temperature, persona, and embedding-selection variants (reported at $r{=}8$ for the latter).
\textit{Correctness noise}, the final-answer mismatch rate in the pool, tracks the no-judge variants.
\textit{Rewrite share}, the fraction of \texttt{wrong} $\to$ \texttt{check} $\to$ \texttt{fix} traces in the pool, tracks the trace-repair variants.

\subsection{Analysis for RQ3}

\paragraph{Audited variants do not surpass RS.}
Fig.~\ref{fig:rq3-panels} shows that each audited protocol moves its targeted intrinsic metric, yet transfer tracks the RS baseline across the method$\times r$ grid.
Response diversity, correctness noise, and correction-signal share each shift under their variants, but the Physics replication yields the same outcome (Tab.~\ref{tab:appendix-rq3-physics}), and averaging each method's gap to RS over the full $r\in\{1,2,4,8,16,32\}$ sweep yields no consistent positive advantage (Appendix~\ref{sec:appendix-training}).

\paragraph{Summary.}
This decoupling between intrinsic shifts and transfer holds within the fixed-source regime tested here; under a different source pool, or with larger teacher or question budgets, the same generation protocols may still contribute on top of RS.
Cross-pair replication with a weaker teacher and a different student again shows no gain over RS (Appendix~\ref{sec:appendix-cross-pair}).

\begin{figure}[t]
    \centering
    \includegraphics[width=0.95\columnwidth]{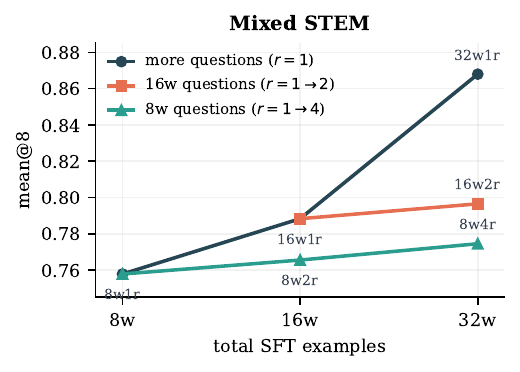}
    \caption{Matched-budget allocation on mixed STEM domains. The `more questions ($r{=}1$)' series scales the seed-question pool; the other two hold the question set at $8$w/$16$w and raise $r$.
    This question-coverage allocation stays strictly above both at every matched budget, and the gap widens with budget.}
    \label{fig:mixed-stem}
\end{figure}

\section{Robustness on a Heterogeneous Source}
\label{sec:mixed-stem}

The main RQ1 and RQ2 experiments use single-domain slices. We relax this single-domain restriction and adopt a heterogeneous source: the STEM set of Nemotron-Post-Training-Dataset-v2~\citep{nvidia2025nvidianemotronnano2}, which mixes STEM domains instead of isolating one domain.
We stress-test both findings on a single matched-budget grid, reading it two ways. Each fixed-pool series traces an FSS response-scaling path (RQ1). The comparison across pool sizes at a fixed total budget is the matched-budget SE-vs-FSS test (RQ2). Here the seeds added at larger $|Q|$ are drawn from a mixture of STEM domains rather than a single one.
We use $T_{\mathrm{DS}}$ (DeepSeek-V3.1) as the teacher and $S_{\mathrm{Q3}}$ (\texttt{Qwen3-4B}) as the student, chosen to keep training cost tractable on this larger heterogeneous pool, and defer source-pool construction, the validation split, and training settings to Appendix~\ref{sec:appendix-heterogeneous}.

Fig.~\ref{fig:mixed-stem} addresses both questions on a single grid.
For RQ1, the two fixed-pool series ($8$w and $16$w) climb only marginally on the $\log_2$ budget axis. This reproduces, on a larger and more heterogeneous source, the diminishing-returns pattern RQ1 established on per-domain slices.
For RQ2, the $r{=}1$ path, which raises real-question coverage, climbs steeply instead. Its advantage over both fixed-pool series widens as the budget grows, and at the matched $32$w total-example budget it lies above both.
On a heterogeneous source, both results hold: the FSS bound persists, and at a fixed total budget, more real seed questions outperform drawing more responses from a smaller seed pool, even when those added seeds span multiple STEM domains.

\section{Discussion}
\label{sec:discussion}

The central distinction is whether a scaling intervention alters the source support or only reallocates budget within it.
Our three questions probe this contrast through different interventions.
Scaling the per-question response budget (Sec.~\ref{sec:rq1}) draws more traces from the fixed source, with diminishing returns on a logarithmic budget axis;
synthesizing additional questions from the existing seeds (Sec.~\ref{sec:rq2}) enlarges the question pool but, at a matched total budget, does not outperform simply sampling more responses;
substituting an alternative synthesis protocol (Sec.~\ref{sec:rq3}) changes how traces are drawn from the same source but does not improve over the RS reference.
Only source expansion escapes this pattern, along both axes of $(Q, T)$: adding real seed questions yields a matched-budget advantage on per-domain slices (Sec.~\ref{sec:rq2}) and on the heterogeneous source (Sec.~\ref{sec:mixed-stem}), and a stronger teacher raises student accuracy at the same sample budget (Sec.~\ref{sec:rq1}).
Under the coverage model (Appendix~\ref{sec:appendix-derivation}), source expansion raises the reachable ceiling, whereas within-source interventions redistribute samples beneath it.
This view explains why FSS scales but is bounded, and why matched-budget comparisons should separate source expansion from within-source synthesis before attributing gains to synthetic-data scaling.

\section{Conclusion}

We separate synthetic-data scaling into \emph{Source Expansion} (SE) and \emph{Fixed-Source Synthesis} (FSS), then isolate FSS by holding $(Q,T)$ fixed.
Across our experiments, FSS follows a bounded response-scaling curve captured by the fixed-source form, while matched-budget gains come from expanding the source rather than reshuffling synthesis within it.
The fixed source therefore provides a controlled setting for evaluating synthesis protocols.

\section*{Limitations}
Prior synthetic-data scaling work focuses primarily on reasoning. Although we cover broader STEM tasks, extending to further domains is left to future work.
Our experiments are computationally expensive. When the teacher is large (for example, DeepSeek-V3.1), generation cost is high, and doubling the sampling budget exceeds our compute envelope. Larger sampling budgets are a natural next step.
Finally, we fix a teacher--student setting across experiments for fair comparison. 
However, synthetic data spans more tasks, and a broader exploration of synthesis methods and deployment scenarios is left to subsequent work.

\section*{Use of AI Assistants}

We primarily use AI assistants to improve and enrich our writing, and to assist with \LaTeX\ formatting.



\bibliography{custom}

\appendix

\section{Appendix}
\label{sec:appendix}

\subsection{Derivation of the FSS response-scaling form}
\label{sec:appendix-derivation}

We derive Eq.~\ref{eq:rq1-hypothesis} from a binary feature-coverage model under four modelling assumptions: binary coverage, linear additivity, (near-)i.i.d.\ exposure, and a heavy-tailed feature-frequency distribution. The heavy-tail assumption is independently motivated by the long-tailed skill-usage frequencies documented by \citet{michaud2023quantization}, who tie precisely such a power-law frequency structure to neural scaling. The empirical fits in Sec.~\ref{sec:rq1} test the resulting functional form in aggregate rather than verify each assumption pointwise.

\paragraph{Setup.}
Fix $(Q,T)$ and let $\mathcal{F}$ be a latent feature space (skills, reasoning patterns, answer templates). For $f\in\mathcal{F}$, define the per-budget-step hit probability
\[
q(f) \;:=\; \Pr[\,\text{one budget step exposes } f\,],
\]
where one budget step corresponds to drawing one response for each question in the fixed pool $Q$ under the sampling protocol (prompt, decoding, filtering), so that $|Q|$ is held constant and absorbed into $q(f)$, and $r$ counts the number of such steps. Increasing the response budget $r$ samples more heavily from this fixed source-induced distribution; it does not change the distribution itself.

\paragraph{Binary coverage and linear additivity.}
We make three modelling assumptions: (i)~\emph{binary coverage}: once $f$ has appeared at least once in the training data, the student can benefit from it at test time; (ii)~\emph{linear additivity}: the test score is a linear combination of per-feature contributions $a(f)\ge 0$; and (iii)~\emph{(near-)i.i.d.\ exposure}: the $r$ draws are treated as independent, each exposing $f$ with probability $q(f)$, so the probability that $f$ is uncovered after $r$ draws is $(1-q(f))^r$ (acceptance filtering and finite-pool sampling make assumption (iii) an approximation rather than exact). Hence
\begin{align*}
P(r) &\;=\; \mathbb{E}_f\!\bigl[a(f)\bigl(1-(1-q(f))^r\bigr)\bigr] \\
     &\;=\; M - \mathbb{E}_f\!\bigl[a(f)(1-q(f))^r\bigr],
\end{align*}
where $M(Q,T) := \mathbb{E}_f[a(f)]$ is the total source contribution mass. The performance attainable from the fixed source is the limit
\[
P_\infty(Q,T) \;:=\; \lim_{r\to\infty} P(r) \;\le\; M,
\]
equal to the contribution mass carried by features with $q(f)>0$, with equality $P_\infty=M$ iff every contributing feature is reachable; this $P_\infty$ is the ceiling fit empirically in Eq.~\ref{eq:rq1-hypothesis}, and $P(r)\le P_\infty\le M$ holds by construction. Rescaling so $M=1$ and absorbing $a(f)$ into the feature measure, so that all expectations below, and the density $g$, are taken under this contribution-weighted, normalized measure, gives
\begin{equation}
P(r) \;=\; 1 - \mathbb{E}_f\bigl[(1-q(f))^r\bigr],
\label{eq:appendix-coverage}
\end{equation}
a change of units, not a new assumption, under which $P_\infty = 1 - \Pr_f[q(f)=0]\le 1$.

\paragraph{Monotonicity, concavity, and the source ceiling.}
From Eq.~\ref{eq:appendix-coverage}, the discrete gain is
\[
P(r{+}1)-P(r) \;=\; \mathbb{E}_f\bigl[q(f)\,(1-q(f))^r\bigr] \;\ge\; 0,
\]
so $P(r)$ is non-decreasing in $r$ (prediction~(a)). For any $f$ with $0<q(f)<1$, the integrand $q(f)(1-q(f))^r$ is strictly decreasing in $r$, hence $P(r{+}2)-P(r{+}1) \le P(r{+}1)-P(r)$: the curve is discretely concave (prediction~(b)). Letting $r\to\infty$ in Eq.~\ref{eq:appendix-coverage} recovers, consistently with the definition above,
\[
P_\infty \;=\; 1 - \Pr_f[q(f)=0] \;\le\; 1,
\]
i.e.\ the reachable ceiling equals the (contribution-weighted) test mass of features the fixed source can ever expose, and attains the normalized bound $1$ only when no contributing feature has $q(f)=0$. Features with $q(f)=0$ cannot be reached by any FSS budget; source expansion changes $(Q,T)$ and can convert some such features to $q(f)>0$, raising the ceiling itself; this is the mechanism behind the SE/FSS separation in Sec.~\ref{sec:rq2}.

\paragraph{Heavy-tail near zero $\Rightarrow$ power law.}
The monotonicity and concavity above are discrete-$r$ statements; for the saturation \emph{rate} we treat $r$ as a continuous variable and study the $r\to\infty$ asymptotic of the gap, of which the discrete differences above are the integer-grid counterpart. Let $g$ denote the density of $q(f)$ on $(0,1)$ under the contribution-weighted, normalized feature measure of Eq.~\ref{eq:appendix-coverage} (point masses at $q=0$ and $q=1$, if present, contribute to $1-P_\infty$ and an $r$-independent constant, respectively). Motivated by the long-tailed skill-usage frequencies documented by \citet{michaud2023quantization}, assume a \emph{pure power-law tail} at zero,
\[
g(q) \;\sim\; c\,q^{\alpha-1} \quad (c>0)\ \text{ as } q\to 0^+,
\]
for some $\alpha>0$ (the constant-coefficient special case of regular variation with index $\alpha-1$; the general case is treated in the remark below). The point mass at $q=0$ contributes $1-P_\infty$ to $\mathbb{E}_f[(1-q)^r]$ at every $r$, and the point mass at $q=1$ contributes zero, so the saturation gap reduces to the integral over the open interval:
\[
P_\infty - P(r) \;=\; \int_{(0,1)} (1-q)^r\,g(q)\,dq.
\]
For large $r$, $(1-q)^r \approx e^{-rq}$ is sharply concentrated near $q=0$, so Watson's lemma (equivalently, the Karamata Tauberian theorem for Laplace transforms) gives
\begin{align*}
P_\infty - P(r)
&\;\sim\; \int_0^\infty e^{-rq}\,c\,q^{\alpha-1}\,dq \\
&\;=\; c\,\Gamma(\alpha)\,r^{-\alpha} \;=:\; A\,r^{-\alpha},
\end{align*}
as $r\to\infty$, giving the form in Eq.~\ref{eq:rq1-hypothesis}. This is an asymptotic equivalence rather than an identity at finite $r$: if the swept budget range already lies in the tail-dominated regime where the equivalence is tight, parameters fit on lower $r$ will extrapolate to higher $r$; whether the regime has been reached at the budgets actually swept is itself an empirical diagnostic (prediction~(c)), which the in-sample-to-out-of-sample test in Sec.~\ref{sec:rq1} probes directly.

\emph{Remark (slowly varying correction).} Under the strictly weaker hypothesis that $g$ is merely regularly varying at zero with index $\alpha-1$, i.e.\ $g(q)=q^{\alpha-1}L(q)$ with $L$ slowly varying, the same Tauberian argument gives $P_\infty-P(r)\sim \Gamma(\alpha)\,r^{-\alpha}\,L(1/r)$: a power law modulated by a slowly varying factor, which collapses to the three-parameter form of Eq.~\ref{eq:rq1-hypothesis} only when $L(1/r)\to\text{const}$. A non-constant $L$ is thus one concrete way the parametric form can hold only approximately even while predictions~(a)--(b) still do; the in-sample fits in Sec.~\ref{sec:rq1} bear on the pure-power case and do not separately identify $L$.

\paragraph{When the heavy-tail step fails.}
The coverage model above implies monotonicity and discrete concavity for any distribution of $q(f)$, but the power-law rate requires mass near $q=0$. If $g$ instead concentrates away from zero, for example in a roughly homogeneous regime $q(f)\approx q$, then $\mathbb{E}_f[(1-q(f))^r]\approx(1-q)^r\approx e^{-qr}$, yielding exponential rather than polynomial saturation. If $g$ consists of well-separated bands, the gap becomes a sum of exponentials with distinct rates. These alternatives would still be monotone and concave, but would not obey the three-parameter form of Eq.~\ref{eq:rq1-hypothesis}; the empirical fits in Sec.~\ref{sec:rq1} therefore test the pure-power tail approximation rather than the coverage model as a whole.

\paragraph{Relation to the rectified scaling law.}
The derived form is, algebraically, the offset-free member of the family of Eq.~\ref{eq:rectified}: writing the accuracy curve as an error rate ($L=1-P$), $P(r)=(1-E)-A\,r^{-\alpha}$ matches Eq.~\ref{eq:rq1-hypothesis} with $P_\infty=1-E$ and the same power, i.e.\ Eq.~\ref{eq:rectified-acc} at $D_\ell{=}0$ with $D$ replaced by $r$. The agreement is one of functional form only, and the two differ where it matters. The fitted floor $E$ of Eq.~\ref{eq:rectified} is a model-capacity and task quantity, whereas $P_\infty$ here is, under (i)--(iii), the contribution mass of features the fixed source can ever expose to the student under a held-constant student and training protocol; variation in $P_\infty$ across fixed-source sweeps is thus interpreted as variation in the source's reachable mass given that fixed student, rather than as a capacity limit revealed by increasing data quantity. The absence of $D_\ell$ is likewise not a fitted simplification but a consequence of conditioning on a fixed source: source expansion, which converts features with $q(f)=0$ to $q(f)>0$ (the mechanism behind the SE/FSS separation in Sec.~\ref{sec:rq2}), moves the process outside this offset-free fixed-source derivation and would re-introduce the additional flexibility associated with $D_\ell$ in Eq.~\ref{eq:rectified}. The external law thus fixes the functional family; the fixed-source mechanism selects its offset-free member and re-interprets the ceiling.

\subsection{Held-out forecast across all pair--domain cells}
\label{sec:appendix-form}

Appendix~\ref{sec:appendix-derivation} selects the three-parameter form $P(r)=P_\infty-A\,r^{-\alpha}$ from the fixed-source derivation. Two empirical checks accompany that selection: 
(i) the held-out forecast of Fig.~\ref{fig:rq1-forecast} extended to all eight pair--domain cells, and 
(ii) a sanity check against trivial extrapolators.

\paragraph{Protocol.}
For each of the eight pair--domain cells (four teacher--student pairs $\times$ two primary domains, Mathematics and Physics), we fit Eq.~\ref{eq:rq1-hypothesis} on $r \le 16$ and forecast $r=32$, following the Physics protocol of Fig.~\ref{fig:rq1-forecast}, under bounds $c \le 1$, $\alpha \ge 0.02$, $A \ge 0$. We report the held-out forecast MAE in percentage points. As trivial extrapolators we use Carry (the $r{=}16$ measurement extended unchanged to $r{=}32$) and LogLin (a straight line in $y$ vs $\log r$ fit on $r \le 16$ and extended to $r{=}32$).

\begin{table}[h]
    \centering
    \small
    \setlength{\tabcolsep}{4pt}
    \begin{tabular}{@{}lc@{\hspace{1.4em}}lc@{}}
        \toprule
        \multicolumn{2}{c}{Mathematics} & \multicolumn{2}{c}{Physics} \\
        \cmidrule(lr){1-2}\cmidrule(lr){3-4}
        Pair & fc & Pair & fc \\
        \midrule
        $T_{DS}{\to}S_{Q7}$  & 0.59 & $T_{DS}{\to}S_{Q7}$  & 0.54 \\
        $T_{DS}{\to}S_{L8}$  & 0.49 & $T_{DS}{\to}S_{L8}$  & 0.76 \\
        $T_{Q72}{\to}S_{Q7}$ & 0.30 & $T_{Q72}{\to}S_{Q7}$ & 0.01 \\
        $T_{Q72}{\to}S_{L8}$ & 0.67 & $T_{Q72}{\to}S_{L8}$ & 0.29 \\
        \midrule
        \textit{mean}        & 0.51 & \textit{mean}        & 0.40 \\
        \bottomrule
    \end{tabular}
    \caption{Forecast MAE at $r{=}32$ (pp) for Eq.~\ref{eq:rq1-hypothesis} fit on $r \le 16$, across all eight pair--domain cells.}
    \label{tab:appendix-form}
\end{table}

\paragraph{Reading.}
Pooled across the eight cells, the held-out forecast MAE is $0.45$ pp for Eq.~\ref{eq:rq1-hypothesis}, against $0.85$ pp for Carry and $0.54$ pp for LogLin. The two trivial extrapolators fail in complementary regimes: Carry degrades on cells where the curve is still rising between $r{=}16$ and $r{=}32$, and LogLin overshoots on cells that have approached saturation. 

\subsection{Synthesis protocols used in the main experiments}

Tab.~\ref{tab:appendix-fss-methods} lists the response-side synthesis protocols audited in RQ3; all variants keep the same seed questions $Q$ and teacher $T$, so each row isolates the effect of the synthesis protocol $p$. The shared backbone is RS: the teacher produces candidate traces per seed question, an LLM judge accepts traces whose final answer matches the gold answer, and the SFT buckets at $r\in\{1,2,4,8,16,32\}$ take the first $r$ accepted traces per prompt. Unless a row states otherwise, the prompt template, teacher endpoint, judge, SFT conversion, student training, and evaluation protocol are unchanged from RS. The correctness judge is Qwen2.5-32B-Instruct decoded at temperature $0.0$ with the prompt shown below; the gold answer is supplied, so the judge checks answer equivalence against the reference rather than solving the problem from scratch. The primary DeepSeek-V3.1 teacher uses up to $8192$ generation tokens, and the generator samples at temperature $1.2$ by default. The question-side protocol used as paradigm 2 in RQ2 is question evolution: additional prompts are synthesized from the same source seeds via Evol-Instruct~\citep{xu2023wizardlm} before RS, so this condition changes the prompt inventory without adding real source questions, and we report it in Fig.~\ref{fig:question-synthesis}. The distinction matters when interpreting the main results: a gain from response-side synthesis reflects trace construction, whereas a gain from question-side synthesis reflects a change in the questions exposed to the student.

\begin{table*}[t]
    \centering
    \footnotesize
    \setlength{\tabcolsep}{3pt}
    \renewcommand{\arraystretch}{1.12}
    \begin{tabular}{p{0.17\textwidth}p{0.29\textwidth}p{0.47\textwidth}}
        \toprule
        Variant & Change relative to RS & Implementation detail \\
        \midrule
        RS & Temperature $1.2$ teacher draws; keep responses & The LLM judge is applied after each teacher trace; prompts that do not supply enough accepted traces are excluded from the complete RS pool used for the corresponding $r$ buckets. \\
        Lower-temperature RS & Draw at temperature $0.6$ & Identical RS pipeline and judge; only the teacher sampling temperature changes. \\
        Lower-temp RS & Draw at temperature $0.001$ & Identical RS pipeline and judge; this setting isolates low decoding entropy. \\
        Embedding-based selection & Select diverse responses from the RS accepted pool, applying the diversity-selection criterion of DEITA~\citep{liu2023-what-makes-good-data} on the response axis & Responses are embedded with \texttt{sentence-transformers/all-MiniLM-L6-v2}; within each prompt, greedy cosine-threshold selection always keeps the first response and admits later responses only when their maximum similarity to the selected set is below a threshold. The threshold is binary-searched separately for each target $r$. \\
        Persona prompting & Condition generation on sampled personas~\citep{ge2024-scaling-synthetic-data-personas} & Personas are randomly sampled from PersonaHub \texttt{elite\_persona} records with a fixed sampler seed and inserted into the conversation, by default as a system message instructing the teacher to answer from that persona while still solving accurately. Judge filtering and SFT conversion are unchanged. \\
        No-judge filtering & Keep raw generated responses without judge filtering, motivated by ~\citet{mukherjee2023orca,xu2025harnessing-negative-signals} & The judge adapter is replaced with an always-pass function that accepts every non-exceptional generated response and marks it as accepted without validation. This variant therefore changes both correctness filtering and, because RS drops prompts without enough accepted traces, prompt coverage. \\
        Prompt-aligned no-judge filtering & Use raw responses while matching the RS prompt subset; no-judge variant~\citep{mukherjee2023orca,xu2025harnessing-negative-signals} with prompt coverage controlled & The raw no-judge pool is filtered to keep the same prompt subset as the RS complete pool, so the comparison varies response filtering while holding prompt coverage fixed. \\
        Repair trace & Convert wrong responses into \texttt{wrong $\to$ check $\to$ fix} traces, instantiating the iterative refinement / critique-then-revise recipes of \citet{madaan2023-self-refine}, \citet{yang2024-supercorrect}, and \citet{kapusuzoglu2025-critique-guided-distillation} & Wrong raw rollouts from the prompt-aligned no-judge pool are paired with one RS-correct trace for the same question. A generator writes an \texttt{ATTEMPT}/\texttt{CHECK}/\texttt{FIX}/\texttt{FINAL ANSWER} trace; generated repairs are judged again, accepted repairs are placed first, and any remaining slots up to $32$ are filled with RS-correct fallback traces. \\
        Revision-only repair & Keep only the revised correct solution from the repair pipeline; ablation that drops the critique trace from \citet{kapusuzoglu2025-critique-guided-distillation} & Uses the same wrong-rollout/reference pairing, re-judging, ordering, and RS-correct fallback rule as repair trace, but the generator outputs only the corrected solution rather than the explicit critique trace. \\
        \bottomrule
    \end{tabular}
    \caption{Synthesis protocols audited in RQ3. All keep $(Q, T)$ fixed and vary only the synthesis protocol $p$.}
    \label{tab:appendix-fss-methods}
\end{table*}

\paragraph{Judge prompt.}
The judge uses the template below; the three placeholders are filled with the source question, gold target, and candidate response.

\begin{tcolorbox}[
  enhanced,
  breakable,
  colback=white,
  colframe=promptBlue,
  width=\columnwidth,
  arc=2mm,
  boxrule=0.5mm,
  title={\normalsize\textbf{Prompt:} Correctness Judge},
  fonttitle=\bfseries\normalsize,
  fontupper=\footnotesize,
]
\begin{Verbatim}[breaklines=true,breakanywhere=true]
You are a helpful assistant who evaluates the correctness and quality of models' outputs. Please as a grading expert, judge whether the final answers given by the candidates below are consistent with the standard answers, that is, whether the candidates answered correctly.

Here are some evaluation criteria:
1. Please refer to the given standard answer. You don't need to re-generate the answer to the question because the standard answer has been given. You only need to judge whether the candidate's answer is consistent with the standard answer according to the form of the question. Don't try to answer the original question. You can assume that the standard answer is definitely correct.
2. Because the candidate's answer may be different from the standard answer in the form of expression, before making a judgment, please understand the question and the standard answer first, and then judge whether the candidate's answer is correct, but be careful not to try to answer the original question.
3. Some answers may contain multiple items, such as multiple-choice questions, multiple-select questions, fill-in-the-blank questions, etc. As long as the answer is the same as the standard answer, it is enough. For multiple-select questions and multiple-blank fill-in-the-blank questions, the candidate needs to answer all the corresponding options or blanks correctly to be considered correct.
4. Some answers may be expressed in different ways, such as some answers may be a mathematical expression, some answers may be a textual description, as long as the meaning expressed is the same. And some formulas are expressed in different ways, but they are equivalent and correct.
5. If the prediction is given with \boxed{{}}, please ignore the \boxed{{}} and only judge whether the candidate's answer is consistent with the standard answer.

Please judge whether the following answers are consistent with the standard answer based on the above criteria. Grade the predicted answer of this new question as one of:
A: CORRECT
B: INCORRECT
Just return the letters "A" or "B", with no text around it.

Here is your task. Simply reply with either CORRECT, INCORRECT. Don't apologize or correct yourself if there was a mistake; we are just trying to grade the answer.

<Original Question Begin>:
{{origin_question}}
<Original Question End>

<Gold Target Begin>:
{{gold_target}}
<Gold Target End>

<Predicted Answer Begin>:
{{predicted_answer}}
<Predicted End>

Judging the correctness of candidates' answers:
\end{Verbatim}
\end{tcolorbox}

\paragraph{Evol-Instruct prompt.}
Question evolution uses DeepSeek-V3.1 as the synthesis model with the generation prompt below.

\begin{tcolorbox}[
  enhanced,
  breakable,
  colback=white,
  colframe=promptBlue,
  width=\columnwidth,
  arc=2mm,
  boxrule=0.5mm,
  title={\normalsize\textbf{Prompt:} Evol-Instruct},
  fonttitle=\bfseries\normalsize,
  fontupper=\footnotesize,
]
\begin{Verbatim}[breaklines=true,breakanywhere=true]
System:
You are an expert question writer for rigorous multiple-choice exams. Generate a new question that is related but not a paraphrase. The new question must be self-contained, have exactly one unambiguous correct option, and remain in the same field/subfield and jurisdictional context implied by the original. Before answering, internally solve the new question yourself and verify that exactly one option is defensible. Return only JSON.

User:
Given this original multiple-choice question:

{seed_prompt}

Create ONE NEW multiple-choice question.

Requirements:
- Same field: {field}
- Same subfield: {subfield}
- Similar difficulty: {difficulty}
- If the original question is tied to a specific jurisdiction, legal system, named law, theorem family, or scientific concept, keep the evolved question in that same scope.
- Exactly {num_options} options
- Exactly one correct answer
- Must NOT be a paraphrase of the original question
- Options should be mutually distinct and plausible distractors
- Do not reuse the original answer text or explanation; create a fresh answer and fresh explanation for the new question
- The distractor options should each be plausible but clearly inferior to the correct answer
- Avoid questions where multiple options can be defended with different assumptions
- The correct answer must be derivable from standard knowledge/reasoning, not hidden assumptions
- The explanation must justify why the chosen option is correct
- `answer_text` must be EXACTLY the text of the correct option, copied verbatim from the options list
- If you are not confident that the question has one unique correct answer, do not improvise; instead generate a simpler but still new question in the same scope

Output format (JSON only):
{
  "question": "...",
  "options": ["...", "...", "...", "..."],
  "answer_letter": "A",
  "answer_text": "...",
  "explanation": "..."
}
\end{Verbatim}
\end{tcolorbox}

\subsection{Training and evaluation details}
\label{sec:appendix-training}

The main experiments share a single SFT and evaluation protocol across methods. We train students for $8$ epochs and evaluate only the final checkpoint. SFT uses AdamW with learning rate $5\times10^{-6}$, minimum learning rate $5\times10^{-7}$ under a warmup-plus-cosine schedule, weight decay $0.1$, global batch size $32$, micro-batch size $1$, maximum sequence length $4096$, and expand-soft globally packed training examples. The primary metric is $\mathrm{mean@8}$: we sample $8$ student responses per test question at inference temperature $0.7$ and judge correctness with the same LLM judge used throughout the study. We report main-table cells with three train seeds whenever the corresponding SFT sets already existed under the repeat protocol: the original run plus two additional train-and-evaluate repeats. Appendix~\ref{sec:appendix-rq3-physics} discusses the reduced Physics coverage in the RQ3 audit and our reasons for not regenerating missing SFT sets. 
For RQ3, we therefore summarize each audited protocol by its mean gap to the RS reference across the full sweep $r\in\{1,2,4,8,16,32\}$. On Mathematics, the eight audited protocols have sweep-mean gaps from $-0.99$pp to $+0.35$pp, with an aggregate mean of $-0.23$pp over all $48$ method--budget points. This supports the fixed-source conclusion that the audited protocol changes give no consistent lift over RS under our protocol, while leaving open narrower interventions or regimes outside this audit.

\subsection{RQ3 audit on Physics}
\label{sec:appendix-rq3-physics}

Tab.~\ref{tab:appendix-rq3-physics} reports the three methods. The qualitative pattern matches Mathematics. Each protocol shifts its intended intrinsic axis on Physics: persona widens pairwise Jaccard distance from $0.48$ to $0.55$; no-judge filtering raises the wrong-response share from $3.3\%$ to $43.2\%$; the prompt-aligned variant raises it to $15.4\%$. The corresponding best-$r$ deltas are small and centered near or below zero. The one slightly positive cell, no-judge filtering at $+0.31_{\pm 0.06}$pp, does not carry over to its prompt-aligned variant ($-0.51_{\pm 0.45}$pp), so the pooled audit gives no support for a method-level Physics improvement either.

\begin{table*}[t]
    \centering
    \footnotesize
    \setlength{\tabcolsep}{4pt}
    \renewcommand{\arraystretch}{1.10}
    \begin{tabular}{l l l l r}
        \toprule
        Method & Intended axis & Intrinsic metric & Intrinsic move & Best-$r$ $\Delta$ (pp), $\mathrm{mean@8}_{\pm\mathrm{std}}$ \\
         & & & (RS$\to$method) & \\
        \midrule
        Persona prompting       & response diversity & pairwise Jaccard dist. & $0.48\!\to\!0.55$ & $-0.84_{\pm 0.41}$ \\
        No-judge filtering      & correctness noise  & wrong-response \%      & $3.3\!\to\!43.2$  & $+0.31_{\pm 0.06}$ \\
        Prompt-aligned no-judge & correctness noise  & wrong-response \%      & $3.3\!\to\!15.4$  & $-0.51_{\pm 0.45}$ \\
        \bottomrule
    \end{tabular}
    \caption{RQ3 audit on Physics for the three methods with replicated runs available. \emph{Intrinsic move} reports the shift from the RS reference to the audited pool; the rightmost column is the best-$r$ $\mathrm{mean@8}_{\pm\mathrm{std}}$ delta in pp against the same-domain RS baseline ($n{=}3$ train seeds).}
    \label{tab:appendix-rq3-physics}
\end{table*}

\subsection{Cross-pair replication of the fixed-source audit}
\label{sec:appendix-cross-pair}

To check that the fixed-source result of Sec.~\ref{sec:rq3} does not hinge on the primary teacher--student pair, we re-run the audit on Mathematics under two pair swaps, a weaker teacher ($T_{\mathrm{Q72}}{\to}S_{\mathrm{Q7}}$) and a second student ($T_{\mathrm{DS}}{\to}S_{\mathrm{L8}}$), with each cell backed by $n{=}3$ train seeds.

\begin{table}[t]
    \centering
    \small
    \setlength{\tabcolsep}{6pt}
    \begin{tabular}{lc}
        \toprule
        Method & mean@8 \\
        \midrule
        \multicolumn{2}{l}{\textit{$T_{\mathrm{DS}}{\to}S_{\mathrm{L8}}$, Mathematics}} \\
        RS                                & $.386_{\pm.006}$ \\
        Lower-temperature RS ($T{=}0.6$)  & $\mathbf{.395}_{\pm.007}$ \\
        Lower-temp RS ($T{=}0.001$)       & $.363_{\pm.009}$ \\
        Embedding-based selection         & $.392_{\pm.006}$ \\
        Persona prompting                 & $.367_{\pm.007}$ \\
        Judge-free filtering              & $.388_{\pm.011}$ \\
        Trace-level repair                & $.388_{\pm.017}$ \\
        \midrule
        \multicolumn{2}{l}{\textit{$T_{\mathrm{DS}}{\to}S_{\mathrm{L8}}$, Physics}} \\
        RS                                & $.327_{\pm.004}$ \\
        Lower-temperature RS ($T{=}0.6$)  & $.327_{\pm.008}$ \\
        Lower-temp RS ($T{=}0.001$)       & $.317_{\pm.009}$ \\
        Embedding-based selection         & $\mathbf{.328}_{\pm.005}$ \\
        Persona prompting                 & $.315_{\pm.010}$ \\
        Judge-free filtering              & $.325_{\pm.009}$ \\
        Trace-level repair                & $.324_{\pm.005}$ \\
        \midrule
        \multicolumn{2}{l}{\textit{$T_{\mathrm{Q72}}{\to}S_{\mathrm{Q7}}$, Mathematics}} \\
        RS                                & $.338_{\pm.010}$ \\
        Lower-temperature RS ($T{=}0.6$)  & $\mathbf{.347}_{\pm.007}$ \\
        Embedding-based selection         & $.330_{\pm.003}$ \\
        Persona prompting                 & $.346_{\pm.006}$ \\
        \bottomrule
    \end{tabular}
    \caption{Fixed-source method audit replicated under teacher--student pair swaps: a second student on both Mathematics and Physics, and a weaker teacher on Mathematics. Each cell reports best-$r$ student $\mathrm{mean@8}$ as $\mathrm{mean}_{\pm\mathrm{std}}$ over $n{=}3$ train seeds; bold marks the strongest cell within each block.}
    \label{tab:cross-pair-methods}
\end{table}

The pattern of Sec.~\ref{sec:rq3} carries across all three swaps (Tab.~\ref{tab:cross-pair-methods}). On Mathematics, lower-temperature RS edges marginally above the RS reference and every other variant lands at or below it; on Physics even that edge disappears, with all variants at or below RS. The fixed-source result thus reproduces under a weaker teacher, a second student family, and a second domain: once $(Q, T)$ is fixed, no audited variant gives a stable improvement over RS.

\subsection{Heterogeneous-source check details}
\label{sec:appendix-heterogeneous}

The robustness check of Sec.~\ref{sec:mixed-stem} draws its source pool from the STEM set of Nemotron-Post-Training-Dataset-v2. From the $355{,}000$ raw examples we take the final answer from \texttt{<answer>} tags or the last boxed expression, yielding $341{,}388$ parsed questions. We keep the parsed order and split these into $324{,}318$ training/source questions and a $1{,}024$-question validation set; the matched-budget grid uses prefixes of the training/source partition as source pools. $T_{\mathrm{DS}}$ is the teacher and $S_{\mathrm{Q3}}$ the student, trained for $5$ SFT epochs; all other settings follow the shared protocol of this appendix.

\end{CJK}  
\end{document}